\documentclass[journal]{IEEEtran}
\usepackage{amsmath,amsfonts}
\usepackage{algorithmic}
\usepackage{algorithm}
\usepackage{array}
\usepackage[caption=false,font=normalsize,labelfont=sf,textfont=sf]{subfig}
\usepackage{textcomp}
\usepackage{stfloats}
\usepackage{url}
\usepackage{verbatim}
\usepackage{graphicx}
\usepackage{cite}

\usepackage{tabularx}
\usepackage{multirow}
\usepackage{makecell}
\usepackage{booktabs}

\begin{document}

\title{DSADiff: Data Structure-Aware Diffusion for Crystal Structure Prediction via Mitigating Complexity- and Imbalance-Induced Biases}

\author{Jiang Xiaobo, Dinghong Lai
        % <-this % stops a space
\thanks{}% <-this % stops a space
\thanks{}}

\markboth{}%
{}

\IEEEpubid{}

\maketitle

\begin{abstract}

Crystal Structure Prediction (CSP) is an important step in accelerating the generation and discovery of new materials. In recent years, diffusion-based CSP methods have made significant progress. However, existing models still suffer from generation bias when facing diverse materials spaces, while systematic approaches for tracing the sources of such performance variations remain limited. To address this issue, we propose LAS (Layered Analysis Strategy), a mechanism analysis framework that identifies crystal complexity and elemental composition imbalance as two key factors affecting generation difficulty and further reveals their relationships with model generation performance. Building on these findings, we further propose DSADiff (Data Structure-Aware Diffusion), a data structure-aware diffusion model that transforms the key factors identified by LAS into a unified generation difficulty representation and uses it to jointly drive feature modeling and training strategy optimization. This enables the model to adapt its representation and generation capability to samples with different data structures and difficulty levels, thereby mitigating learning bias across samples. Experiments on three CSP benchmark datasets, including Carbon-24, demonstrate that DSADiff consistently improves both Match Rate and RMSE, with a maximum Match Rate improvement of 9\%, surpassing existing state-of-the-art methods and achieving new state-of-the-art performance. This work deepens the understanding of diffusion-based CSP models from the perspective of generation bias and provides a new direction for designing less biased generative models for complex materials spaces.

\end{abstract}


\section{Introduction}
\IEEEPARstart{C}{rystal} materials constitute the foundation of many modern material systems and have broad applications in fields such as energy storage and electronic devices \cite{ref1}. CSP aims to generate three-dimensional crystal structures solely from chemical compositions \cite{ref2}, making it a key task in inverse materials design \cite{ref3}. Traditional CSP approaches typically rely on experimental trial-and-error or high-throughput computational methods \cite{ref4}. Although these methods can produce reliable crystal structures, they are often computationally expensive.

In contrast, deep generative models can directly generate candidate crystal structures, significantly reducing the reliance on costly energy calculations and manual search strategies \cite{ref5}. Among them, diffusion models have become the dominant approach for CSP owing to their strong capability in modeling complex data distributions \cite{ref6}. Representative methods, such as DiffCSP \cite{ref44} and TGDMat \cite{ref45}, have achieved remarkable performance on multiple public benchmark datasets.

Although existing diffusion models have substantially advanced the performance of CSP and previous studies have recognized the influence of training data on generation quality, investigations into performance variations have primarily focused on factors such as dataset size and chemical space coverage \cite{ref27}. In contrast, the effects of crystal complexity and elemental composition imbalance on the learning process of diffusion models remain largely unexplored. In practice, crystal datasets and material systems exhibit considerable differences in compositional diversity, crystal size, and other intrinsic characteristics, resulting in varying levels of generation difficulty. Moreover, the distribution of elemental compositions in materials databases is inherently imbalanced, leading to different learning difficulties between high-frequency and low-frequency compositions.

Such variations in crystal complexity and elemental composition imbalance may introduce systematic generation bias into diffusion-based CSP models. For example, representative diffusion models exhibit substantial differences in Match Rate and RMSE across different crystal datasets \cite{ref45}. However, existing CSP studies have primarily attributed these performance differences to model architecture or the joint modeling of atomic coordinates and lattice parameters \cite{ref44}, while data-related analyses have remained largely limited to dataset size and chemical space coverage \cite{ref27}. However, existing studies still lack systematic analytical methods for tracing the sources of such performance variations. Meanwhile, several fundamental questions remain unanswered: Is the generation difficulty of crystal structures inherently determined by the characteristics of crystal data? Is there an intrinsic relationship between data heterogeneity and the performance of diffusion-based CSP models? How can such information be exploited to mitigate the generation difficulty of diffusion-based CSP models?

To address the above issues, we propose a LAS method to systematically investigate the effects of crystal complexity and elemental composition imbalance on the performance of diffusion-based CSP models from a data-centric perspective. By establishing the relationship between crystal data characteristics and model performance, LAS reveals the underlying mechanisms responsible for performance variations across different types of crystal data, providing a new perspective for understanding the learning behavior of diffusion-based CSP models in diverse crystal spaces.

Building upon these findings, we further propose DSADiff, a data structure-aware diffusion model for crystal structure generation. DSADiff jointly encodes crystal complexity and elemental composition imbalance into a unified difficulty descriptor, which is used to dynamically guide the model learning process. On the one hand, a difficulty-aware feature adaptation mechanism enables the model to accommodate the representation requirements of crystals with different levels of complexity. On the other hand, an adaptive training strategy alleviates the learning bias caused by elemental composition imbalance.

To validate the effectiveness of DSADiff, extensive experiments are conducted on multiple public CSP benchmark datasets with diverse crystal complexity and elemental composition imbalance characteristics. The experimental results demonstrate that DSADiff consistently improves both Match Rate and RMSE, outperforming existing baseline methods while exhibiting strong generalization across different datasets.

The main contributions of this work are summarized as follows:

\begin{itemize}
\item We introduce the concept of crystal complexity, which characterizes the structural properties of crystal samples from multiple aspects affecting crystal generation difficulty, including compositional diversity, crystal size, and local geometric complexity.

\item We propose a LAS mechanism analysis strategy to systematically investigate the effects of crystal complexity and elemental composition imbalance on the performance of diffusion-based CSP models, establishing the relationship between crystal data characteristics and generation performance.

\item Based on the proposed mechanism analysis, we develop DSADiff, a data structure-aware diffusion model for crystal structure generation. By jointly modeling generation difficulty through a unified difficulty descriptor and introducing adaptive learning strategies, DSADiff effectively mitigates the learning bias caused by crystal complexity and elemental composition imbalance.

\item Extensive experiments on CSP benchmark datasets with diverse crystal complexity and elemental composition imbalance, including Carbon-24, demonstrate that DSADiff consistently outperforms existing state-of-the-art methods and achieves new state-of-the-art performance.

\end{itemize}

\section{Related work}
\subsection{Generative model}

The rapid development of generative artificial intelligence has introduced a new paradigm for inverse materials design \cite{ref7}. By learning the statistical distributions of existing materials, generative models can directly produce candidate materials with desired properties, thereby substantially reducing the computational cost associated with conventional trial-and-error approaches and high-throughput screening \cite{ref8}. In recent years, a variety of generative models, including variational autoencoders (VAEs) \cite{ref9}, generative adversarial networks (GANs) \cite{ref10}, diffusion models \cite{ref11}, and large language models (LLMs) \cite{ref12}, have been successfully applied to materials design, significantly accelerating the inverse design of diverse materials \cite{ref13}. Representative examples include GNoME developed by Google DeepMind \cite{ref14} and MatterGen proposed by Microsoft \cite{ref15}.

Early studies on materials generation primarily relied on VAEs and GANs. VAEs learn continuous latent representations of material structures, enabling continuous optimization of both chemical compositions and crystal structures. Court et al. \cite{ref16} proposed a VAE-based generative deep representation learning framework capable of generating three-dimensional crystal structures while simultaneously predicting their target properties. Ren et al. \cite{ref17} introduced a general inverse design framework that generated a large number of crystal structures across multiple design tasks. In contrast, GANs focus on modeling complex material distributions. Kim et al. \cite{ref18} combined GAN-based generative modeling with high-throughput virtual screening for photoanode materials, leading to the discovery of 23 novel crystal structures exhibiting reasonable thermodynamic stability and desirable band gaps. Zhao et al. \cite{ref19} proposed CubicGAN, a GAN-based framework for large-scale cubic crystal generation, which produced a diverse set of hypothetical materials with previously unseen structural prototypes. Collectively, these studies demonstrated that generative models are capable of effectively exploring unexplored chemical spaces and have established a promising new paradigm for accelerated materials discovery.

In recent years, diffusion models have gradually become the dominant paradigm for crystal structure generation owing to their stable training dynamics and outstanding capability for modeling complex data distributions. Luo et al. \cite{ref20} proposed SyMat, a materials generation framework that integrates variational autoencoders with diffusion models to capture the physical symmetries inherent in periodic crystal structures. Zhong et al. \cite{ref21} introduced CHGGen, a conditional generative framework that leverages embedding techniques to address the challenge of generating symmetry-consistent crystal structures. Klipfel et al. \cite{ref22} developed a diffusion model that jointly generates lattice parameters and atomic coordinates using an equivariant graph neural network, enabling the discovery of crystal structures with targeted chemical properties. Yang et al. \cite{ref23} proposed UniMat, a unified crystal representation framework coupled with diffusion probabilistic modeling, which facilitates high-fidelity crystal generation for larger and more complex chemical systems.

Meanwhile, LLMs have also emerged as a promising approach for crystal generation. Yan et al. \cite{ref24} proposed Mat2Seq, a sequence transformation method that assigns each crystal a unique and canonical sequence representation, thereby resolving the ambiguity introduced by directly modeling crystallographic information files (CIFs). Antunes et al. \cite{ref25} further developed CrystaLLM, which is capable of generating physically plausible crystal structures for a wide range of inorganic compounds that were absent from the training data, demonstrating the potential of LLMs to learn effective representations of crystal chemistry.

Overall, the development of existing generative models has primarily focused on enhancing representation learning and generative capability through improvements in network architectures, material representations, and conditional generation strategies. However, relatively little attention has been paid to how the intrinsic complexity of crystal structures and the imbalance in elemental composition distributions affect the learning dynamics and generation performance of these models.

\subsection{Crystal structure prediction}

Crystals constitute the fundamental structural form of a vast majority of materials, and their three-dimensional atomic arrangements determine the resulting physical and chemical properties, playing a pivotal role in applications ranging from drug discovery to heterogeneous catalysis \cite{ref26}. CSP aims to determine the stable crystal structure and identify potential polymorphs of a material solely from its chemical composition, making it a fundamental task in computational materials science \cite{ref27}. Traditional CSP methods primarily rely on DFT for energy evaluation \cite{ref28}, combined with optimization algorithms such as genetic algorithms \cite{ref29}, particle swarm optimization \cite{ref30}, and evolutionary algorithms \cite{ref31} to search for local minima on the potential energy surface. These approaches have achieved remarkable success in predicting inorganic compounds, including superconducting hydrides and electronic materials \cite{ref32}. However, the extensive DFT calculations required incur prohibitively high computational costs and present a persistent trade-off between efficiency and accuracy \cite{ref33}, thereby limiting the applicability of conventional DFT-based CSP methods to relatively small systems or restricted configurational spaces \cite{ref34}.

With the rapid expansion of crystal structure databases and advances in artificial intelligence, deep generative approaches for CSP have emerged as a promising alternative \cite{ref35}. Among them, diffusion models, inspired by nonequilibrium thermodynamics \cite{ref36}, establish a mapping between complex data distributions and simple prior distributions through forward and reverse Markov diffusion processes \cite{ref37}. Their outstanding performance in image generation \cite{ref38}, molecular conformation generation \cite{ref39}, protein structure prediction \cite{ref40}, and protein docking \cite{ref41} has attracted widespread attention, further promoting the adoption of diffusion models for CSP.

To accommodate the intrinsic characteristics of crystalline materials, including periodicity, symmetry, and physical constraints, extensive efforts have been devoted to adapting diffusion models for crystal structure generation. Gebauer et al. \cite{ref42} proposed P-cG-SchNet, a conditional generative framework for three-dimensional molecular structures that demonstrated the feasibility of sampling novel configurations from conditional distributions, laying the foundation for extending diffusion-based generation to periodic systems. Xie et al. \cite{ref43} introduced the CDVAE, which progressively refines atomic coordinates toward low-energy configurations through a diffusion process while simultaneously updating atomic species to satisfy local bonding constraints. Moreover, CDVAE explicitly preserves permutation, translation, rotation, and periodic invariance throughout the generation process.

Building upon CDVAE, Jiao et al. \cite{ref44} proposed DiffCSP, which employs a periodic E(3)-equivariant denoising architecture to jointly generate lattice parameters and atomic coordinates. By representing crystal structures using fractional rather than Cartesian coordinates, DiffCSP substantially improves CSP performance. More recently, Das et al. \cite{ref45} developed TGDMat, which further incorporates textual conditions into the diffusion process by jointly modeling natural language descriptions and crystal structures. This text-guided generation framework achieves state-of-the-art performance across multiple benchmark datasets while simultaneously reducing computational cost.

Although these studies have significantly advanced diffusion-based CSP, their improvements have primarily focused on model-centric aspects, such as network architecture design, incorporation of physical priors, and conditional information integration. Meanwhile, existing studies have recognized that the training data influence generation quality, yet most investigations have been limited to factors such as dataset size and chemical space coverage \cite{ref27}. Whether in CDVAE, DiffCSP, or TGDMat, relatively little attention has been paid to two fundamental characteristics that are pervasive in real-world crystal datasets: crystal complexity and elemental composition imbalance. In practice, crystal samples exhibit substantial heterogeneity in compositional diversity, structural scale, and local geometric complexity, while their elemental compositions are typically distributed in a highly imbalanced manner. These characteristics jointly affect the model optimization process during training and are likely to contribute to the considerable performance variations observed across different CSP benchmark datasets.

\section{Preliminaries}

\subsection{Datasets}
Following previous studies \cite{ref45}, we select three representative benchmark datasets for CSP to systematically evaluate the generative capability of models under varying levels of structural complexity and material distribution characteristics.

The Perov-5 dataset \cite{ref46,ref47} contains 18,928 perovskite structures, where all samples share the same structural prototype but differ in chemical composition. Each unit cell consists of five atoms, and the structures exhibit relatively concentrated configurations and regular crystal geometries. Therefore, this dataset is widely used to evaluate the capability of models in learning perovskite structural patterns and accurately reconstructing similar crystal structures.

The Carbon-24 dataset \cite{ref48} comprises 10,153 carbon-based crystal materials, with the number of atoms per unit cell ranging from 6 to 24. Unlike Perov-5, samples in Carbon-24 share the same chemical composition but possess diverse structural prototypes. Consequently, this dataset exhibits higher diversity in terms of atomic-scale configurations and local structural environments, making it suitable for evaluating the ability of models to capture complex structural variations beyond compositional differences.

The MP-20 dataset \cite{ref43} is derived from the Materials Project database \cite{ref49} and consists of 45,231 stable inorganic materials covering a broad range of experimentally feasible crystal structures. Unlike Perov-5 and Carbon-24, samples in MP-20 contain diverse compositions and structural prototypes, with up to 20 atoms per unit cell. Due to its substantial compositional and structural diversity, MP-20 has become one of the most representative benchmarks for CSP. It is commonly adopted to comprehensively evaluate the generation quality and stability of models in large-scale and heterogeneous inorganic material spaces.

By conducting experiments on these three complementary datasets, we can comprehensively assess model performance and robustness across different scenarios, including simple structural generation, complex structure reconstruction, and large-scale realistic material generation.

\subsection{Evaluation metrics}

During the evaluation stage, we follow the standard protocol commonly adopted in CSP tasks \cite{ref43}. For each ground-truth structure in the test set, we generate k = 1 candidate structure with the same chemical composition and perform structural matching between the generated and reference structures using StructureMatcher implemented in pymatgen \cite{ref50}. The matching criteria are configured with $\mathit{stol} = 0.5,\ \mathit{angle\_tol} = 10,\ \mathit{ltol} = 0.3$. A test sample is considered successfully predicted if at least one generated candidate matches the corresponding ground-truth structure. Accordingly, the Match Rate (MR) is defined as the proportion of successfully matched samples among all test samples:

\begin{equation}
\label{1}
\mathrm{MR}=\frac{1}{N}\sum_{i=1}^{N}m_{i},\quad m_{i}=\mathbb{I}\left(\exists j\in\{1,\ldots,k\},d_{ij}\neq\mathrm{None}\right)
\end{equation}

where $N$ denotes the total number of test samples, $d_{ij}$ represents the RMS distance between the $i$-th ground-truth structure and the $j$-th generated candidate structure, and $m_{i}$ is a sample-level matching indicator.

In addition, RMSE is used to further quantify the geometric reconstruction error of successfully matched samples. For each test sample, the candidate structure with the minimum geometric deviation from the ground-truth structure is first selected from all valid generated candidates:

\begin{equation}
\label{2}
d _ {i} ^ {*} = \min _ {j: d _ {i j} \neq \mathrm {N o n e}} d _ {i j}
\end{equation}

Then, the overall RMSE is calculated by averaging the errors over only the successfully matched samples.

\begin{equation}
\label{3}
\mathrm {RMSE} = \frac {1}{N_{\mathrm {match}}} \sum_ {i=1}^{N}m_{i} \tilde {d}_{i}^{*}, \quad N _ {\mathrm {match}} = \sum_{i = 1}^{N}m_{i}
\end{equation}

where $\tilde{d}_{i}^{*}$ denotes the normalized RMS distance. Following the normalization scheme adopted in this work, it is calculated as:

\begin{equation}
\label{4}
\tilde {d}_{i}^{*} = \frac {d_{i}^{*}}{3pV_{i}/N_{i}}
\end{equation}

where $V_i$ represents the lattice volume and $N_i$ denotes the number of atoms in the corresponding unit cell.

Furthermore, to ensure consistent and fair comparisons among different approaches, both the baseline models and DSADiff model are evaluated under the same setting of $K = 1$. Under this setting, each test sample generates only a single candidate structure. Therefore, Match Rate directly reflects the capability of a model to recover the ground-truth structure under single-shot generation, while RMSE further characterizes the geometric deviation between the generated structure and the reference structure.

\subsection{Model overview}
To systematically investigate the effects of crystal complexity and elemental composition imbalance on the performance of crystal generation, and to validate the effectiveness of the proposed method, we conduct our study based on two representative diffusion-based crystal generation models.

\begin{figure}[!t]
\centering
\includegraphics[width=2.5in]{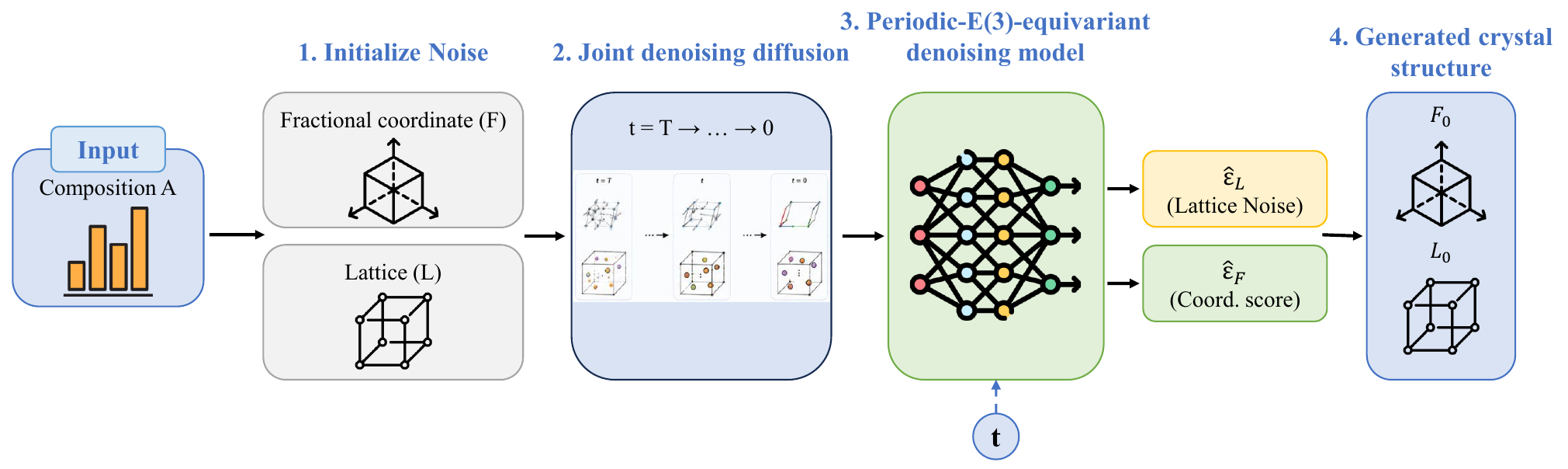}
\caption{Simplified architecture of the DiffCSP model}
\label{fig1}
\end{figure}

DiffCSP is a representative periodic equivariant diffusion model that learns the structural distribution of stable crystals. It employs a periodic E(3)-equivariant denoising network to jointly model lattice parameters and atomic positions during the generation process, thereby providing a more accurate representation of the overall crystal geometry. In addition, DiffCSP represents atomic positions using fractional coordinates, which are naturally consistent with the periodic boundary conditions of crystal structures. A schematic overview of the model is shown in Fig. \ref{fig1}.

\begin{figure}[!t]
\centering
\includegraphics[width=2.5in]{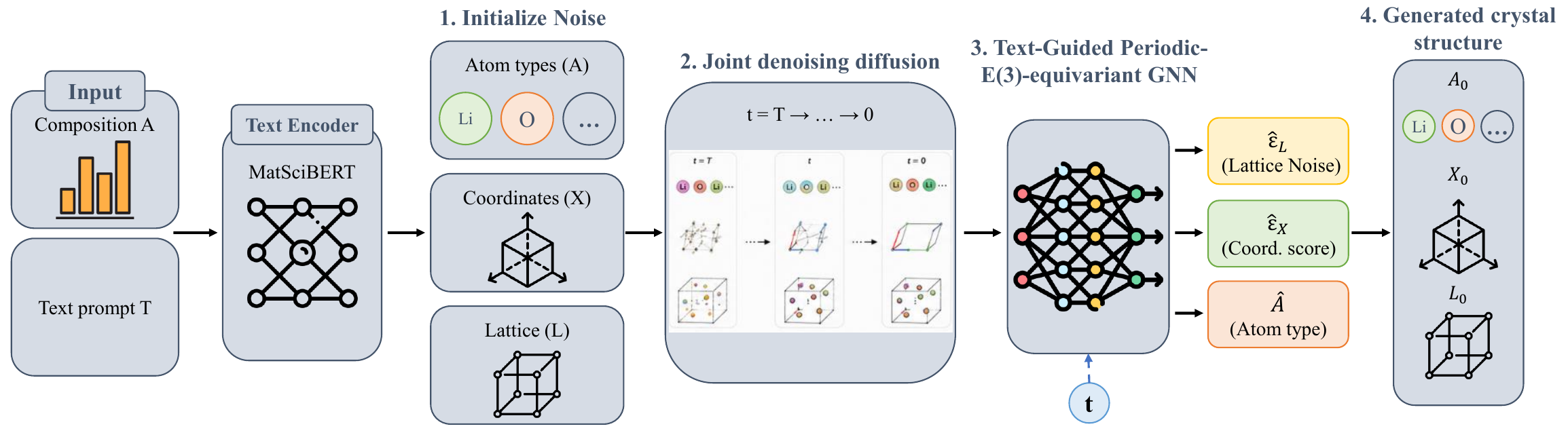}
\caption{Simplified architecture of the TGDMat model}
\label{fig2}
\end{figure}

TGDMat further extends this framework by introducing a text-guided mechanism into periodic crystal generation to compensate for the limited capability of conventional diffusion models in capturing global structural knowledge. Specifically, TGDMat also adopts a periodic E(3)-equivariant graph neural network as the denoising backbone and jointly generates atomic species, fractional coordinates, and lattice structures within a unified diffusion framework, enabling end-to-end modeling of three-dimensional periodic crystals. A schematic overview of the model is shown in Fig. \ref{fig2}.

\section{Mechanism Analysis LAS}

Given the substantial differences among the three datasets in terms of chemical composition space, structural diversity, and sample distribution, we propose a Layered Analysis Strategy (LAS) to systematically investigate the underlying mechanisms across different types of datasets, thereby improving the robustness and interpretability of the conclusions \cite{ref51}.

Specifically, Perov-5 represents a typical perovskite dataset with a fixed structural prototype but diverse chemical compositions. Although the crystal framework and spatial arrangement patterns remain relatively consistent, frequent elemental substitutions introduce compositional variations. In contrast, Carbon-24 consists of materials with an identical chemical composition but different structural prototypes, where the chemical elements are fixed while variations in atomic arrangements, lattice configurations, and symmetry characteristics are more prominent. Compared with these two datasets, MP-20 covers a broader range of realistic inorganic materials, involving diverse elemental compositions, crystal systems, space groups, and lattice configurations. Consequently, MP-20 exhibits stronger sample heterogeneity and a wider compositional distribution, making it a more suitable primary benchmark for analyzing the relationship between data complexity, compositional imbalance, and model generation performance.

Based on this consideration, we select MP-20 as the primary dataset for mechanism analysis and employ correlation analysis, SHAP-based feature importance analysis, and OLS regression analysis with controlled variables to systematically investigate the potential relationships between crystal complexity, elemental composition imbalance, and crystal generation performance. The analyses conducted on Perov-5 and Carbon-24 are further used as complementary validation to examine the generalizability of the observations. Moreover, to eliminate the potential influence of text guidance, all mechanism analyses are performed using DiffCSP, a diffusion model without text conditioning.

\subsection{Crystal complexity}

To quantify the amount of effective constraint information provided by an individual crystal sample in the CSP task, we define a crystal complexity metric $\rho$. This metric comprehensively characterizes the information richness contained in a crystal structure from three perspectives: compositional diversity, structural scale, and local geometric complexity. Intuitively, a larger $\rho$ indicates that the crystal contains richer chemical and geometric constraints, providing the model with more discriminative information during the denoising and structure reconstruction processes. Conversely, a smaller $\rho$ suggests a sparser structural representation with weaker constraints, which may introduce greater ambiguity and difficulty in the generation process.

Specifically, $\rho$ is formulated as a weighted combination of three interpretable components:

\begin{equation}
\label{5}
\begin{split}
\rho &= w_{1}H_{\mathrm {norm}} + w_{2}S_{\mathrm {norm}} + w_{3}G_{\mathrm {loc}}, \\
w_1 &= 0.40, w_2 = 0.35, w_3 = 0.25, \sum_{i} w_{i} = 1
\end{split}
\end{equation}

where $H_{\text{norm}}$ denotes the normalized elemental entropy, which characterizes the compositional diversity of a crystal structure:

\begin{equation}
\label{6}
\begin{split}
H = - \sum_ {z \in \mathcal {Z}} p_{z} \log p_{z}, \quad p_{z} = \frac {n_{z}}{N}, \\
H_{\mathrm {norm}} = \frac {H}{\log|\mathcal {Z}|} \in [0, 1]
\end{split}
\end{equation}

where $Z$ represents the set of distinct elements contained in the sample, and $n_{z}$ denotes the number of atoms of element $z$, $N=\sum_{z} n_{z}$. A larger value of $H_{\text{norm}}$ indicates that the structure contains more diverse elemental species with a more balanced distribution, thereby providing richer chemical constraint information.

The second component, $S_{\text{norm}}$, represents the structural scale factor, which captures the influence of crystal size on complexity:

\begin{equation}
\label{7}
S_{\mathrm{norm}} = \frac {\log (N+1)}{\log \left(N_{\mathrm {r e f}} + 1\right)} \in [0, 1]
\end{equation}

where $N_{\text{ref}} = 200$ is the reference upper bound. The logarithmic normalization is adopted to mitigate the linear dominance of large structures, allowing the complexity metric to increase sublinearly with the number of atoms, which better reflects the relationship between structural scale and generation difficulty.

The third component, $G_{\text{loc}}$, measures the local geometric complexity. Specifically, we calculate the minimum-image pairwise distances $d_{ij}$ between atoms based on their fractional coordinates. To reduce computational overhead, at most 64 atoms are retained when calculating this term. Let:

\begin{equation}
\label{8}
\mu = \operatorname {mean} \left(d_{i j}\right), \quad \sigma = \operatorname {std} \left(d_{i j}\right)
\end{equation}

Then, $G_{\text{loc}}$ is defined as:

\begin{equation}
\label{9}
G_{\mathrm {loc}} = \frac {1}{3} \operatorname {clip} \left(\frac {\sigma}{\mu}, 0, 3\right) \in [0, 1]
\end{equation}

This component characterizes the dispersion of local distance distributions. When interatomic distances become more heterogeneous and local geometric environments become more complex, the ratio $\sigma/\mu$ increases, indicating that the structure contains more non-trivial geometric patterns and is therefore more difficult to recover from noisy representations.

The weights are empirically assigned as $\omega_1 = 0.40$, $\omega_2 = 0.35$, and $\omega_3 = 0.25$, based on the physical relevance of each component to the effective structural constraints available for diffusion-based reconstruction. Although the three components characterize complementary aspects of crystal complexity, their contributions to the available constraint information are not equivalent.

Elemental entropy directly reflects the diversity and balance of chemical compositions, which provides fundamental chemical constraints for identifying feasible structures. Therefore, it is assigned the highest weight. The structural scale term captures the influence of atomic number and spatial degrees of freedom, which affects the complexity of the reconstruction space. However, its contribution mainly reflects the expansion of structural degrees of freedom rather than intrinsic discriminative information, and thus receives a slightly lower weight. The local geometric complexity term characterizes fine-grained variations in atomic arrangements and provides important information for local denoising and structural refinement. Nevertheless, this term is more sensitive to sampling strategies, unit-cell representations, and distance statistics, which may introduce additional variance unrelated to intrinsic generation difficulty. Therefore, a relatively conservative weight is assigned to mitigate potential noise amplification.

The above weighting scheme is determined according to domain knowledge rather than optimized on evaluation performance, avoiding potential data leakage and ensuring the generalizability of the proposed metric. Through the weighted aggregation of these three interpretable components, $\rho$ provides a stable and comparable measurement of the structural information available for crystal generation while maintaining physical interpretability across heterogeneous material datasets.

To systematically investigate the impact of crystal complexity on model performance, we conduct a global correlation analysis between crystal complexity and the Match Rate and RMSE metrics using only the MP-20 dataset. This design choice aims to avoid potential statistical instability or biased conclusions caused by the highly concentrated complexity distributions in the Perov-5 and Carbon-24 datasets. As shown in Table \ref{table1}, crystal complexity exhibits a significant negative correlation with Match Rate and a significant positive correlation with RMSE, with both correlations achieving statistical significance ($p < 0.05$). These results indicate that increasing crystal complexity leads to greater difficulty in structural recovery during the denoising process of diffusion models, resulting in lower Match Rate and larger RMSE. Therefore, crystal complexity can be considered an important factor influencing the performance of diffusion-based CSP models.

\begin{table}[!t]
\caption{Correlation analysis on MP-20 dataset\label{table1}}
\centering
\setlength{\tabcolsep}{4pt}
\begin{tabular}{ccccc}
\hline
MP-20 & pearson\_r & pearson\_p & spearman\_r & spearman\_p \\
\hline
Matched & -0.1722 & $3.88 \times 10^{-61}$ & -0.2656 & $6.63 \times 10^{-146}$ \\
RMSE & 0.1843 & $1.81 \times 10^{-35}$ & 0.3911 & $1.88 \times 10^{-163}$ \\
\hline
\end{tabular}
\end{table}

\begin{figure}[!t]
\centering
\includegraphics[width=2.5in]{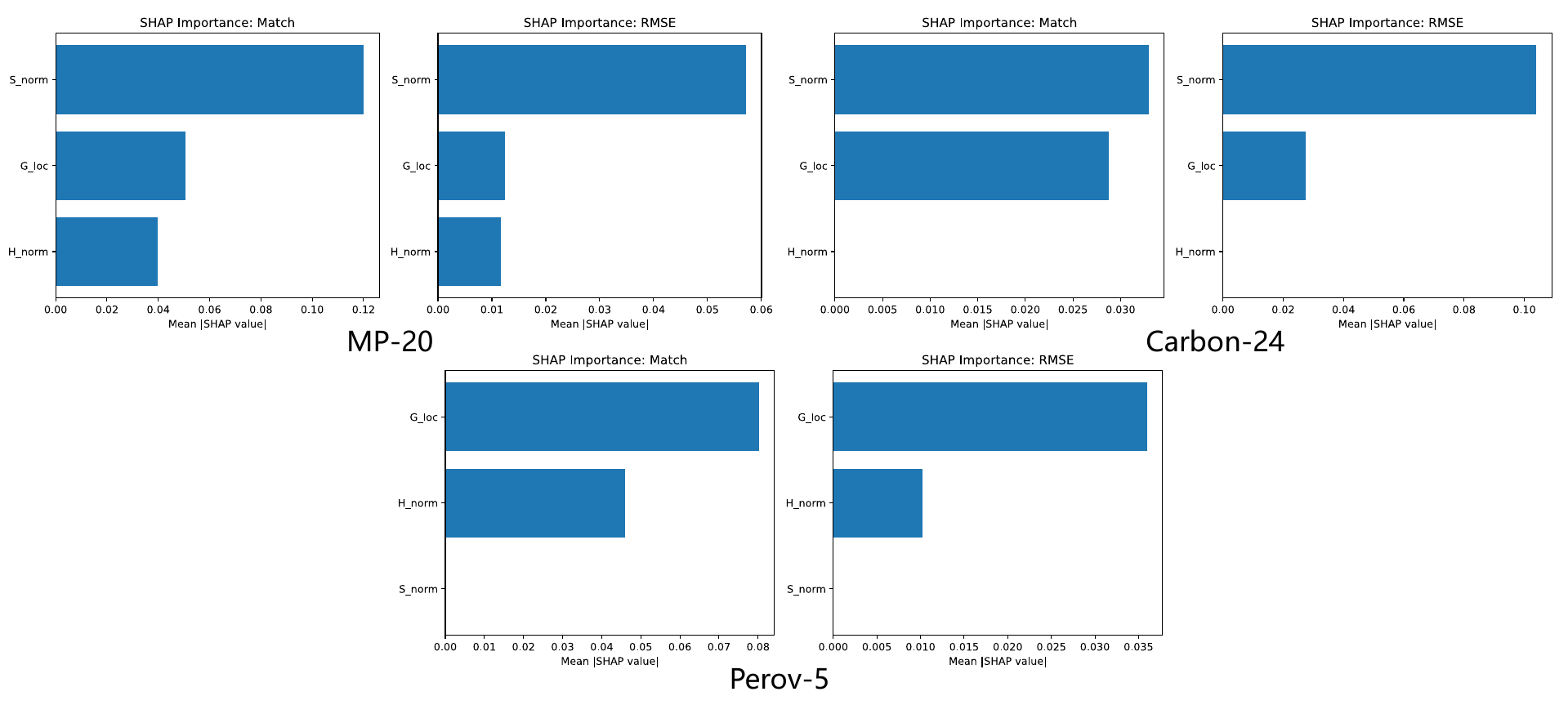}
\caption{SHAP analysis of complexity features}
\label{fig3}
\end{figure}

To further validate this observation, we perform SHAP-based feature importance analysis on the individual components of crystal complexity across the three datasets as complementary evidence. The results are presented in Fig. \ref{fig3}. For the MP-20 dataset, compositional diversity, structural scale, and local geometric complexity all contribute to variations in Match Rate and RMSE, suggesting that multiple sources of crystal complexity jointly affect model performance in realistic inorganic material spaces. In contrast, all samples in the Carbon-24 dataset consist solely of carbon atoms, resulting in identical elemental composition across samples. Consequently, the marginal contribution of $H_{\text{norm}}$ naturally approaches zero. Similarly, Perov-5 contains materials with a shared structural prototype, leading to limited variations in structural scale and framework characteristics, and thus a relatively constrained contribution from $S_{\text{norm}}$.

The consistent observations across the three benchmark datasets demonstrate that the constituent factors of crystal complexity influence the performance of diffusion-based CSP models, although the relative contributions of individual components vary depending on the underlying data characteristics. Overall, these results suggest that crystal complexity affects the strength of training signals, the denoising trajectory, and the final structural reconstruction quality.

These findings provide direct motivation for the subsequent development of a complexity-aware sample-level calibration strategy. By explicitly identifying structurally challenging samples and enhancing their learning signals, the proposed approach has the potential to improve reconstruction performance and generalization capability across diverse material systems.

\subsection{Elemental composition imbalance}

To quantify the impact of distribution-level elemental composition imbalance on crystal generation performance, we analyze the occurrence frequency of elemental compositions in the training set. Specifically, compositions are represented by their elemental count signatures. For example, $\mathrm{SiO}_2$ and $\mathrm{Si}_2\mathrm{O}_3$ share the same elemental composition signature $8\mid14$ and are therefore treated as the same composition category in our analysis.

To characterize the generation behavior of samples with different composition frequencies, we divide the occurrence counts of elemental compositions in the training set into five fixed-frequency bins: $1$, $2$–$3$, $4$–$7$, $8$–$15$, and $>15$. A stratified comparison is then conducted across the three benchmark datasets. This binning strategy approximately follows a logarithmic scale, which preserves fine-grained distinctions among low-frequency compositions while avoiding the limitations of quantile-based binning, where extremely rare and frequent compositions may be grouped together, and equal-width binning, which may excessively compress the long-tail region. Therefore, this strategy is more suitable for investigating how frequency variation under imbalanced composition distributions affects generation quality. This design is also consistent with common practices in long-tailed learning, where frequency groups are typically divided using approximately exponential intervals (e.g., $2^{0}$, $2^{1}$, $2^{2}$, $2^{3}$, and $>2^{3}$) to characterize distributional differences among samples.

\begin{figure}[!t]
\centering
\includegraphics[width=2.5in]{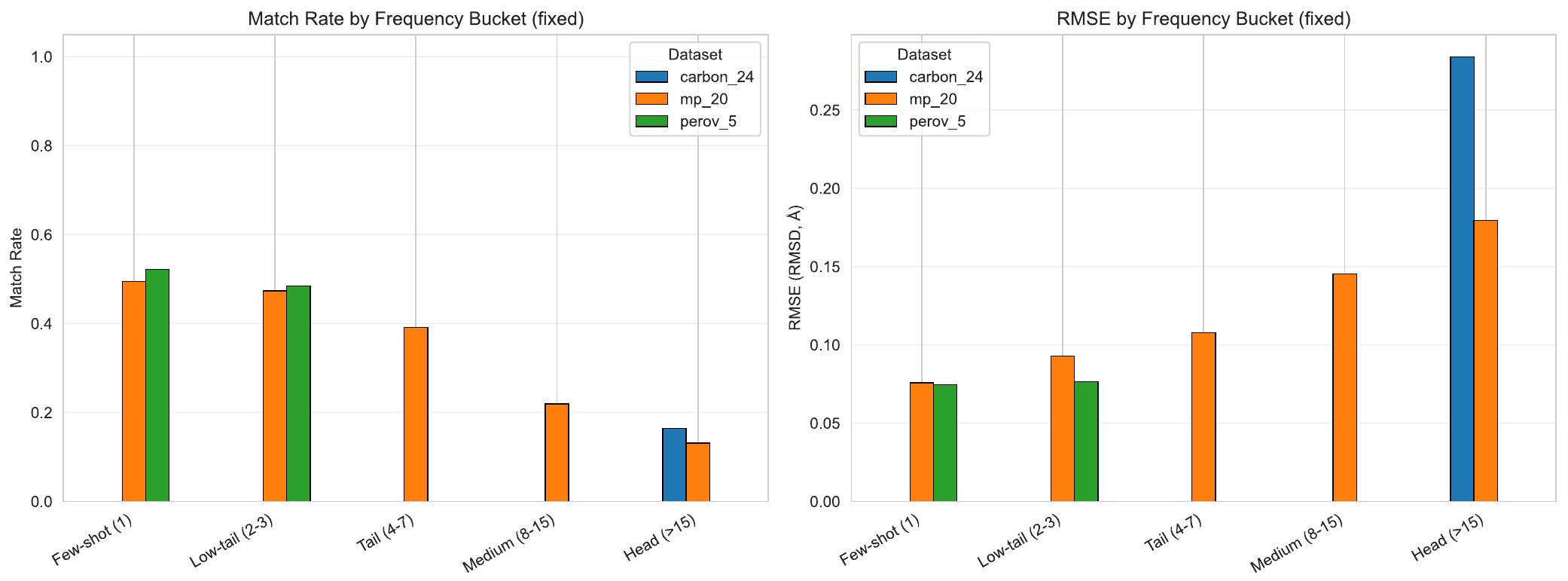}
\caption{Analysis of elemental composition imbalance}
\label{fig4}
\end{figure}

As shown in Fig. \ref{fig4}, MP-20 exhibits the most complete frequency distribution range. Its Match Rate shows a clear decreasing trend as elemental composition frequency increases, while RMSE increases correspondingly, indicating that structures associated with frequently occurring elemental compositions are not necessarily easier to generate and may even exhibit degraded reconstruction quality. Perov-5 demonstrates a similar tendency; however, most compositions occur only 1–3 times, with very limited samples in high-frequency groups, resulting in insufficient frequency variation and reduced statistical robustness. In contrast, all samples in Carbon-24 belong to the same composition-frequency category due to its single-element nature, and therefore it lacks sufficient compositional frequency variation for identifying frequency-related effects.

Overall, the cross-dataset observations indicate that the relationship between generation performance and elemental composition frequency is not merely coincidental, but is closely associated with the distributional imbalance of elemental compositions within the dataset.

To further eliminate potential confounding effects introduced by intrinsic crystal characteristics, we perform an OLS regression analysis with controlled variables at the elemental composition level on the MP-20 dataset. Specifically, the average Match Rate and average RMSE of each elemental composition are used as dependent variables, while $\log(1+\mathit{freq})$ is adopted as the independent variable. The number of elemental species ($n_{\text{elem}}$) and the number of atomic sites ($n_{\text{sites}}$) are included as control variables. In addition, Huber–White heteroscedasticity-robust standard errors (HC3) are employed to alleviate potential estimation bias caused by unequal sample sizes among composition groups.

The logarithmic transformation $\log(1+\mathit{freq})$ is adopted instead of the raw frequency values because composition frequencies exhibit a highly imbalanced distribution. Directly using raw frequencies may allow a small number of highly frequent compositions to dominate the regression estimation, whereas logarithmic transformation reduces the influence of extreme values and better captures the diminishing marginal effect of increasing elemental composition frequency.

The regression results are presented in Table 4.2. After controlling for crystal-related factors, $\log(1+\mathit{freq})$ remains a significant negative predictor of Match Rate ($\beta=-0.1061,\ p\ll0.01$) and a significant positive predictor of RMSE ($\beta=0.0144,\ p<0.05$). These results indicate that elemental composition frequency itself is an important factor associated with generation performance, rather than merely reflecting variations caused by intrinsic differences among crystal structures.

Combining the stratified analysis with the controlled regression results, we further infer that, in material databases, a single elemental composition often corresponds to multiple crystal prototypes or multimodal structural distributions. As elemental composition frequency increases, the model is required to simultaneously capture a broader range of possible structural configurations, thereby increasing the difficulty of target structure identification and leading to larger reconstruction errors. In contrast, low-frequency compositions are often associated with more specific structural patterns and relatively clearer learning objectives, resulting in more stable generation behavior. These findings demonstrate that elemental composition imbalance in material datasets affects the multimodal learning difficulty of diffusion-based CSP models, degrades crystal structure generation quality, and provides direct motivation for the subsequent development of a distribution-aware structured scheduling strategy for addressing compositional imbalance.

\begin{table}[!t]
\caption{Huber-White regression analysis between log(1+freq) and generation performance metrics\label{table2}}
\centering
\setlength{\tabcolsep}{4pt}
\begin{tabularx}{\linewidth}{cccccX}
\hline
MP-20 & Predictor & Controls & Coefficient ($\beta$) & P-value \\
\hline
Match Rate & $\log(1+\mathit{freq})$ & $n_{\text{elem}},\, n_{\text{sites}}$ & $-0.1061$ & $1.99 \times 10^{-18}$ \\
RMSE & $\log(1+\mathit{freq})$ & $n_{\text{elem}},\, n_{\text{sites}}$ & $0.0144$ & $0.038$ \\
\hline
\end{tabularx}
\end{table}

\section{DSADiff model}

The mechanism analysis reveals that the performance variations of diffusion-based CSP models across different datasets are determined not only by the model architecture itself but also by the intrinsic characteristics of crystal data. Specifically, crystal complexity determines the level of constraints required during the diffusion denoising process, whereas elemental composition imbalance affects the opportunities for different compositional categories to participate in model optimization. Together, these two factors lead to insufficient representation capability for structurally complex crystals and elementally imbalanced compositions.

Existing diffusion-based CSP models largely overlook intrinsic crystal characteristics, such as compositional diversity, crystal size, and local geometric complexity, resulting in a mismatch with the underlying distribution of crystal data. This mismatch is a major cause of the performance variation observed across different datasets. Furthermore, the mechanism analysis indicates that simply increasing network depth, enlarging model capacity, or modifying the diffusion architecture cannot fundamentally mitigate the learning bias introduced by data heterogeneity.

Motivated by these findings, we propose DSADiff, a data structure-aware diffusion model for crystal structure generation. DSADiff employs crystal complexity and elemental composition imbalance, the two key factors identified by the mechanism analysis, as unified control signals and establishes a closed-loop framework consisting of diagnosis, conditional representation, and training optimization. Specifically, DSADiff encodes data heterogeneity into a unified difficulty descriptor and explicitly incorporates the key data characteristics affecting crystal generation difficulty during training. This enables the model to establish the relationship between data heterogeneity and the feature learning process, thereby achieving data-aware representation learning and optimization while effectively mitigating the learning bias caused by data heterogeneity.

In the diagnosis layer, DSADiff constructs a crystal complexity descriptor and an elemental composition imbalance descriptor to characterize the generation difficulty of each sample. Specifically, the crystal complexity descriptor consists of normalized elemental entropy, crystal size, and local geometric complexity, which jointly quantify the amount of effective structural constraints available during the diffusion denoising process. The elemental composition imbalance descriptor is computed based on the occurrence frequency of each elemental composition in the training set, reflecting the imbalance of the data distribution. Together, these two descriptors provide a unified representation of sample generation difficulty.

In the Conditional representation layer, the unified difficulty descriptor is mapped into conditional information and injected into the intermediate feature representations of the diffusion network. Specifically, a conditional affine transformation is adopted to generate sample-specific scaling and shifting parameters, enabling conditional reconstruction of the intermediate features according to the sample difficulty. In this way, the model dynamically reallocates its representation capacity according to crystal complexity and elemental composition imbalance, allowing structurally complex crystals and elementally imbalanced compositions to obtain more informative feature representations, thereby improving the overall generation performance.

In the training layer, the same difficulty descriptor is further incorporated into the learning process as an optimization signal, guiding representation learning at both the sample and distribution levels. On the one hand, structural constraints are dynamically adjusted according to crystal complexity, enabling the optimization process to remain responsive to high-difficulty crystals while preventing the model from converging to suboptimal solutions dominated by easier structures. On the other hand, the knowledge acquisition process is dynamically organized according to elemental composition imbalance, where the range of training samples is progressively adjusted throughout the training process. This gradually shifts the learning focus from common elemental compositions to underrepresented ones, achieving progressive coverage of the composition space and enhancing the model's representation capability for elementally imbalanced compositions.

\begin{figure}[!t]
\centering
\includegraphics[width=2.5in]{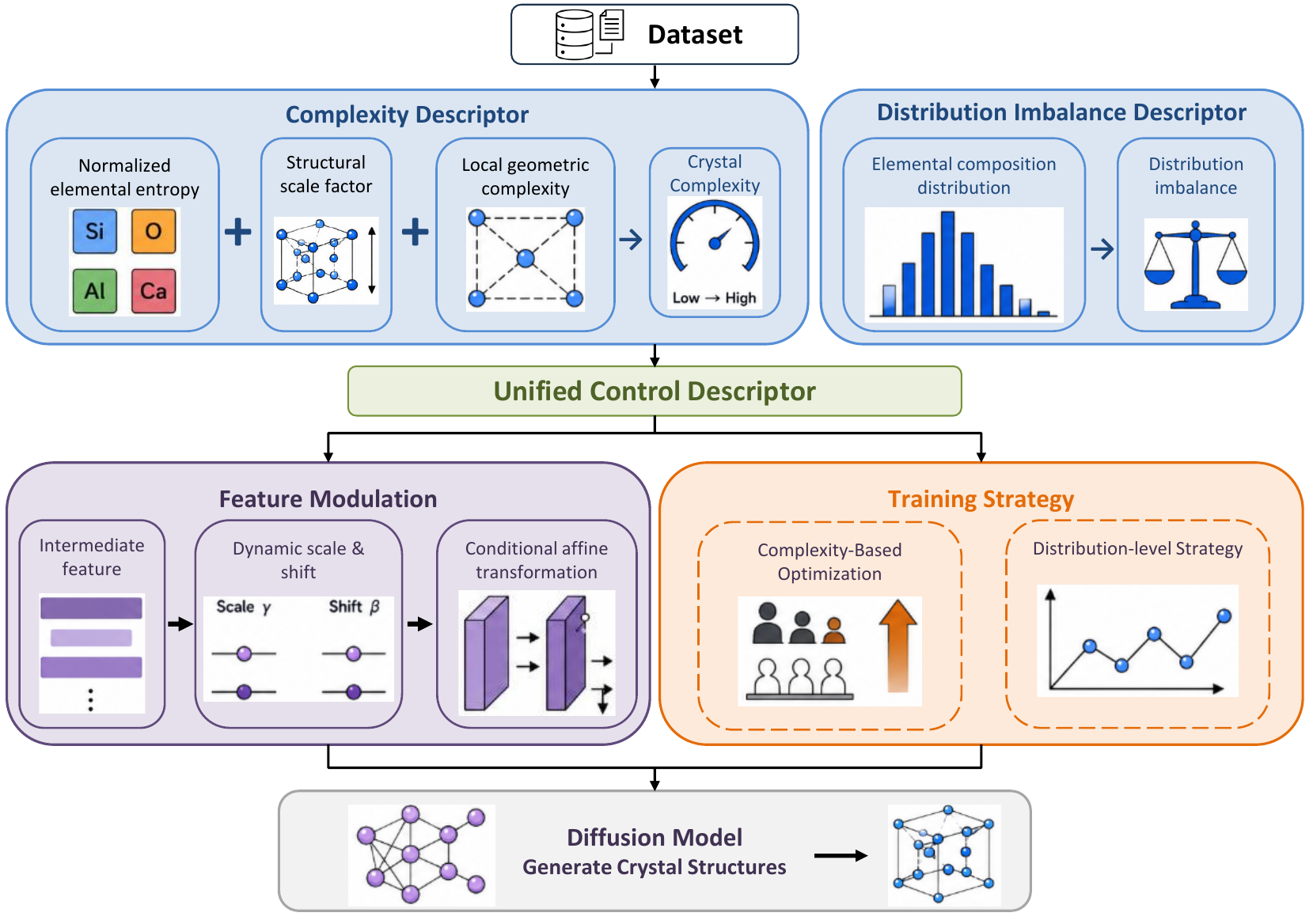}
\caption{Overall architecture of DSADiff model}
\label{fig5}
\end{figure}

Overall, DSADiff is a data structure-aware diffusion model designed based on the findings of the proposed mechanism analysis. The diagnosis layer identifies crystal complexity and elemental composition imbalance, the Conditional representation layer transforms these data characteristics into dynamic conditions for feature representation, and the training layer further mitigates learning bias from the perspective of optimization. Together, these three layers establish a complete learning pipeline spanning data characterization, feature representation, and training optimization, enabling the model to better adapt to crystal samples with different generation difficulties and improve the generation performance on structurally complex crystals and elementally imbalanced compositions. The diffusion generation module of DSADiff follows the architecture of TGDMat (Long) \cite{ref45}. The overall architecture of DSADiff is illustrated in Fig. \ref{fig5}.

\section{Results and discussion}
\subsection{Performance validation and generalization analysis}

To evaluate the effectiveness and generalizability of DSADiff, comparative experiments were conducted on multiple benchmark CSP datasets covering diverse chemical spaces and data scales. This experimental design aims to assess whether the proposed model can consistently improve CSP performance across datasets with different characteristics. The experimental results, summarized in Table \ref{table3}, reveal substantial performance variations among diffusion-based CSP models when only a single candidate structure is generated for each test sample ($K=1$). Building upon these baselines, DSADiff consistently reduces performance variations across different datasets and further improves the reconstruction quality of structurally challenging crystal samples.

Among the baseline methods, P-cG-SchNet, CDVAE, and DiffCSP represent three major generative paradigms for CSP, namely autoregressive generation, variational autoencoders, and diffusion models, respectively. Compared with earlier approaches, DiffCSP achieves superior performance across all four benchmark datasets, demonstrating the stronger distribution modeling capability of diffusion models for CSP. Building upon DiffCSP, TGDMat further incorporates text guidance into the diffusion process and outperforms non-text-guided diffusion models on three of the benchmark datasets, indicating that high-level material prior knowledge can effectively strengthen structural constraints during generation. Nevertheless, it is important to note that neither DiffCSP nor TGDMat explicitly models crystal complexity or elemental composition imbalance, despite their substantial influence on crystal generation performance.

Compared with the baseline methods, DSADiff consistently achieves superior performance across all benchmark datasets in terms of both Match Rate and RMSE, demonstrating its effectiveness in improving crystal structure generation quality. These results indicate that DSADiff effectively mitigates the learning bias associated with crystal complexity and elemental composition imbalance by explicitly incorporating complexity-aware and imbalance-aware optimization into the learning process. This suggests that adapting the optimization process to the intrinsic characteristics of crystal data is an effective strategy for enhancing the robustness and generalization capability of diffusion-based CSP models.

It is worth noting that DSADiff achieves substantial performance improvements over the re-implemented baseline models on the Perov-5, Carbon-24, and MP-20 benchmark datasets. Since differences in implementation details and experimental environments may lead to discrepancies in absolute performance, TGDMat and DiffCSP were re-implemented under a unified experimental setting, and all comparisons were conducted against these consistently reproduced baselines rather than the results reported in the original publications. This evaluation protocol ensures a fair assessment of the performance gains achieved by DSADiff. The consistent improvements observed across multiple benchmark datasets provide strong empirical evidence that crystal complexity and elemental composition imbalance are two key factors limiting the performance of diffusion-based CSP models. More importantly, the results demonstrate that explicitly modeling these factors and incorporating them into the learning process constitute an effective approach for improving crystal structure generation performance.

\begin{table*}[!t]
\centering
\caption{Summary of Experiment Results. TGDMat and DiffCSP are reimplemented under our experimental environment and settings.}
\label{table3}
\footnotesize 
\setlength{\tabcolsep}{4pt} 
\begin{tabular}{c c *{6}{c}}
\toprule
\multirow{2}{*}{Method} & \multirow{2}{*}{\#Samples} & \multicolumn{2}{c}{MP-20} & \multicolumn{2}{c}{Carbon-24} & \multicolumn{2}{c}{Perov-5} \\
\cmidrule(lr){3-4} \cmidrule(lr){5-6} \cmidrule(lr){7-8}
& & Match Rate $\uparrow$ & RMSE $\downarrow$ & Match Rate $\uparrow$ & RMSE $\downarrow$ & Match Rate $\uparrow$ & RMSE $\downarrow$ \\
\midrule
P-cG-SchNet & \multirow{5}{*}{$K=1$}
& 15.39 & 0.3762 & 17.29 & 0.3846 & 48.22 & 0.4179 \\
CDVAE &
& 33.90 & 0.1045 & 17.09 & 0.2969 & 45.31 & 0.1138 \\
DiffCSP &
& 45.47 & 0.0729 & 16.45 & 0.2841 & 51.33 & 0.0763 \\
TGDMat(Long) &
& 52.84 & 0.0944 & 43.60 & 0.3567 & 91.12 & 0.0203 \\
DSADiff (Ours) &
& \textbf{58.09} & \textbf{0.0688} & \textbf{52.61} & \textbf{0.2127} & \textbf{93.18} & \textbf{0.0173} \\
\bottomrule
\end{tabular}
\end{table*}

\subsection{Ablation analysis}

To evaluate the individual contributions of the two functional layers in DSADiff, namely the conditional representation layer and the training layer, and to further investigate whether the performance improvements arise from explicitly addressing data characteristics rather than merely introducing additional model components, we conduct ablation studies from two complementary perspectives: module effectiveness and mechanism-oriented analysis. The former examines the independent and complementary contributions of the conditional representation layer and the training layer through component-wise ablation. The latter combines grouped evaluations to further verify whether DSADiff can specifically improve model performance by addressing the two key data characteristics identified in the mechanism analysis, namely crystal complexity and elemental composition imbalance.

For the module effectiveness study, Carbon-24 is selected as the benchmark dataset. Under the same training configuration and the setting where only one candidate structure is generated for each material ($K=1$), four experimental settings are compared: (1) DSADiff (Base), (2) the complete DSADiff with both the conditional representation layer and the training layer enabled, (3) DSADiff with only the conditional representation layer, and (4) DSADiff with only the training layer. The corresponding results are reported in Table \ref{table4}.

\begin{table}[!t]
\centering
\caption{Ablation study of modules in the DSADiff model}
\label{table4}
\footnotesize
\setlength{\tabcolsep}{6pt}
\begin{tabular}{c c c c c c}
\toprule
Dataset & Method & Conditional & Training & \makecell{Match\\Rate} & RMSE \\
\midrule
\multirow{5}{*}{Carbon-24}
& \multirow{5}{*}{DSADiff} & $\times$ & $\times$ & 43.60 & 0.3567 \\
\cmidrule{3-6}
& & $\checkmark$ & $\times$ & 51.53 & 0.2260 \\
\cmidrule{3-6}
& & $\times$ & $\checkmark$ & 46.85 & 0.3127 \\
\cmidrule{3-6}
& & $\checkmark$ & $\checkmark$ & 52.61 & 0.2127 \\
\bottomrule
\end{tabular}
\end{table}

The ablation results demonstrate that the complete DSADiff model achieves the best performance in terms of both Match Rate and RMSE. Retaining either the conditional representation layer or the training layer individually still leads to better performance than the baseline model, although neither variant matches the performance of the complete model. These results confirm that both components make meaningful and complementary contributions to the overall performance improvement. Combined with the findings of the mechanism analysis, it can be further inferred that the conditional representation layer primarily enhances the model's ability to represent crystal samples with different generation difficulties through adaptive feature representation, whereas the training layer alleviates the learning bias caused by variations in sample difficulty and elemental composition imbalance by adaptively adjusting the training process.

For the grouped analysis of crystal complexity and elemental composition imbalance, the MP-20 dataset, which covers a wide range of chemical compositions and crystal structures, is selected as the benchmark. According to the defined crystal complexity, the dataset is divided into three groups, namely Low, Mid, and High, and the corresponding results are reported in Table \ref{table5}. Both the baseline model and DSADiff exhibit a consistent performance trend across the three complexity groups: as crystal complexity increases, Match Rate decreases while RMSE correspondingly increases, indicating that structurally complex crystals present greater generation difficulty. This observation is consistent with the conclusions of the mechanism analysis, further confirming that crystal complexity is a key factor influencing the performance of diffusion-based CSP models.

After incorporating DSADiff, performance is consistently improved across all three complexity groups. In particular, the Mid and High complexity groups achieve more substantial improvements in both Match Rate and RMSE, whereas the gains for the Low complexity group are relatively modest because the baseline model already achieves strong performance on these comparatively simple samples. These results indicate that the proposed crystal complexity-aware sample-level optimization strategy is particularly effective for challenging crystal structures, improving both the generation success rate and the structural reconstruction accuracy of high-complexity samples, thereby mitigating the performance degradation associated with increasing generation difficulty. Overall, these findings demonstrate that adapting the learning process according to crystal complexity enhances the model's ability to generalize across crystal structures with different levels of generation difficulty.

\begin{table}[!t]
\centering
\caption{Grouped comparison based on crystal complexity}
\label{table5}
\footnotesize
\setlength{\tabcolsep}{5pt}
\begin{tabular}{c *{6}{c}}
\toprule
\multirow{3}{*}{Method} 
& \multicolumn{2}{c}{Low} & \multicolumn{2}{c}{Mid} & \multicolumn{2}{c}{High} \\
\cmidrule(lr){2-3} \cmidrule(lr){4-5} \cmidrule(lr){6-7}
& \makecell{Match\\Rate} & RMSE 
& \makecell{Match\\Rate} & RMSE 
& \makecell{Match\\Rate} & RMSE \\
\midrule
DSADiff(Base) & 78.31 & 0.0494 & 50.51 & 0.1227 & 37.53 & 0.2016 \\
DSADiff       & 82.18 & 0.0339 & 58.93 & 0.0829 & 44.36 & 0.1426 \\
\bottomrule
\end{tabular}
\end{table}

To evaluate the impact of elemental composition imbalance, the test samples were grouped according to the occurrence frequency of their elemental compositions in the training set, and the performance of the baseline model and DSADiff was compared across different frequency groups. The results are presented in Table 6.2.2. The results show that DSADiff improves both evaluation metrics across all groups, indicating that the proposed distribution-level strategy effectively enhances the model's ability to learn elementally imbalanced compositions and improves its generalization to such compositions.

From a group-wise perspective, DSADiff achieves higher Match Rate and lower RMSE across all frequency groups, with substantially larger improvements for low-frequency compositions than for high-frequency ones. Although the Head group exhibits only a modest improvement in Match Rate, it still achieves a considerable reduction in RMSE. This observation suggests that, for high-frequency compositions, the baseline model has already learned the underlying compositional patterns effectively due to their sufficient representation in the training set, leaving limited room for further improvement in structure matching. Nevertheless, DSADiff continues to improve the prediction accuracy of lattice parameters and atomic positions. In contrast, for low-frequency compositions, DSADiff not only significantly increases the success rate of crystal structure generation but also improves the overall quality of the generated structures. These results further validate the effectiveness of the proposed strategy for addressing elemental composition imbalance.

\begin{table}[!t]
\centering
\caption{Grouped comparison based on elemental composition imbalance}
\label{table6}
\footnotesize
\setlength{\tabcolsep}{6pt}
\begin{tabular}{c c c c}
\toprule
\multirow{2}{*}{Group} & \multirow{2}{*}{Metric} & \multicolumn{2}{c}{Method} \\
\cmidrule(lr){3-4}
& & DSADiff(Base) & DSADiff \\
\midrule
\multirow{2}{*}{Zero(0)}
& Match Rate & 62.42 & 67.25 \\
& RMSE & 0.0776 & 0.0533 \\
\midrule
\multirow{2}{*}{Few-shot (1)}
& Match Rate & 58.06 & 65.88 \\
& RMSE & 0.1045 & 0.0731 \\
\midrule
\multirow{2}{*}{Low-tail (2-3)}
& Match Rate & 59.66 & 67.24 \\
& RMSE & 0.1098 & 0.0829 \\
\midrule
\multirow{2}{*}{Tail (4-7)}
& Match Rate & 53.42 & 61.27 \\
& RMSE & 0.1391 & 0.1048 \\
\midrule
\multirow{2}{*}{Medium (8-15)}
& Match Rate & 37.45 & 40.00 \\
& RMSE & 0.1766 & 0.1188 \\
\midrule
\multirow{2}{*}{Head ($>$15)}
& Match Rate & 21.97 & 22.82 \\
& RMSE & 0.1906 & 0.1354 \\
\bottomrule
\end{tabular}
\end{table}

Overall, the observed performance degradation in crystal diffusion-based generation is primarily driven by crystal complexity and elemental composition imbalance. The Conditional representation layer and the training layer provide targeted interventions for these two data characteristics from the perspectives of feature representation and optimization, respectively. Their joint modeling through a unified difficulty descriptor forms a complete mitigation framework, which provides the underlying basis for the performance advantages of DSADiff. The ablation results further verify the effectiveness of each layer in DSADiff.

The above experimental results demonstrate that crystal complexity and elemental composition imbalance, as identified by the proposed mechanism analysis, are the primary factors affecting the performance of diffusion-based CSP models. Accordingly, DSADiff introduces two complementary pathways, namely the Conditional representation layer and the training layer, to address these two data characteristics through feature representation and training optimization, respectively, while jointly modeling them with a unified difficulty descriptor. Specifically, the Conditional representation layer enhances feature representation, whereas the training layer improves the optimization process. Therefore, the performance gains of DSADiff arise not from a simple extension of the model architecture, but from the explicit modeling of data characteristics and targeted interventions. These results further validate the proposed mechanism analysis and demonstrate that DSADiff effectively mitigates the learning bias caused by data heterogeneity, thereby improving the generation performance of diffusion-based CSP models on structurally complex crystals and elementally imbalanced compositions.

\section{Conclusion}

This work systematically reveals the underlying mechanism by which crystal complexity and elemental composition imbalance induce systematic generation bias in diffusion-based CSP models from the perspective of the intrinsic distribution and structural characteristics of crystal data, and accordingly proposes DSADiff, a data structure-aware diffusion model. Our findings demonstrate that the performance bottleneck of diffusion-based CSP models is determined not only by the model architecture itself but also, more fundamentally, by intrinsic data characteristics, including elemental diversity, crystal size, and local geometric complexity. These findings provide a new data-centric perspective for understanding the behavior of generative AI in crystal structure generation.

Based on the proposed LAS mechanism analysis strategy, we further reveal how data characteristics lead to generation bias. Specifically, increasing crystal complexity introduces stronger geometric constraints and greater uncertainty during the diffusion denoising process, resulting in lower Match Rate and higher RMSE, whereas elemental composition imbalance causes biased learning across different compositional spaces. To address these challenges, DSADiff explicitly incorporates complexity priors and composition imbalance representations into the optimization process of diffusion models, enabling adaptive learning based on sample generation difficulty. As a result, it effectively mitigates the systematic bias toward structurally complex crystals and elementally imbalanced compositions.

Extensive experiments on three public CSP benchmark datasets with distinct data distributions, including Carbon-24, demonstrate that DSADiff consistently outperforms existing state-of-the-art methods in terms of both Match Rate and RMSE. In particular, DSADiff achieves an improvement of up to 9\% in Match Rate while exhibiting strong generalization and robustness across different data distributions. Furthermore, experiments grouped by crystal complexity and elemental composition imbalance show that the proposed method consistently improves CSP performance. These results are highly consistent with the findings of the proposed LAS mechanism analysis and further validate the effectiveness of explicitly modeling data characteristics and adaptive optimization strategies.

Overall, this work not only presents a data-centric framework for analyzing and improving diffusion-based CSP models but also provides a new perspective for designing unbiased generative models in complex materials spaces. In future work, we will extend DSADiff to more complex chemical systems, such as high-entropy materials, and investigate property-guided inverse materials design to further advance the application of artificial intelligence in real-world materials discovery.



\bibliographystyle{IEEEtran} 
\bibliography{citation.bib}

\vfill

\end{document}